\pdfoutput=1

\documentclass[11pt]{article}

\usepackage{ACL2023}

\usepackage{times}
\usepackage{latexsym}

\usepackage[T1]{fontenc}

\usepackage[utf8]{inputenc}

\usepackage{microtype}

\usepackage{inconsolata}

\usepackage{graphicx}
\graphicspath{{figures/}}
\usepackage{booktabs}
\usepackage{amsmath}
\usepackage{amssymb}
\usepackage{booktabs}
\usepackage[para,online,flushleft]{threeparttable}
\usepackage{caption}
\usepackage{subcaption}
\usepackage[normalem]{ulem}
\usepackage{colortbl}
\usepackage{enumitem}%

\title{mCPT at SemEval-2023 Task 3: Multilingual Label-Aware Contrastive Pre-Training of Transformers for Few- and Zero-shot Framing Detection}

\author{
  Markus Reiter-Haas\thanks{\enspace equal contribution}\enspace\thanks{\enspace corresponding author} , 
  Alexander Ertl\footnotemark[1] , 
  Kevin Innerebner, 
  Elisabeth Lex \\
  Graz University of Technology, Institute of Interactive Systems and Data Science \\
  Sandgasse 36/III, 8010, Graz, Austria  \\
  \texttt{reiter-haas@tugraz.at, ertl@student.tugraz.at}\\ \texttt{innerebner@student.tugraz.at, elisabeth.lex@tugraz.at} 
}

\begin{document}
\maketitle
\begin{abstract}

This paper presents the winning system for the zero-shot Spanish framing detection task, which also achieves competitive places in eight additional languages. The challenge of the framing detection task lies in identifying a set of 14 frames when only a few or zero samples are available, i.e., a multilingual multi-label few- or zero-shot setting. Our developed solution employs a pre-training procedure based on multilingual Transformers using a label-aware contrastive loss function. In addition to describing the system, we perform an embedding space analysis and ablation study to demonstrate how our pre-training procedure supports framing detection to advance computational framing analysis.

\end{abstract}

\section{Introduction}

Approaches for computational framing detection are diverse~\cite{ali-hassan-2022-survey}, as the framing concept itself is often just casually defined~\cite{entman1993framing}. Consequently, framing detection is challenging on its own, but also suffers from a lack of sufficient data~\cite{kwak2020systematic}, especially in multilingual settings. The SemEval 2023 Task 3 Subtask 2~\cite{semeval2023task3} aims at predicting 14 distinct media frames~\cite{boydstun2013identifying} present within news articles in $9$ languages. 
Due to label imbalances, as a result of the high dimension of the label space compared to the number of samples, traditional paradigms, e.g., per-label binary classification, do not apply well to the given setting without adaptions~\cite{tarekegn2021review}.

\begin{figure}[ht]
    \centering
    \includegraphics[width=.95\linewidth]{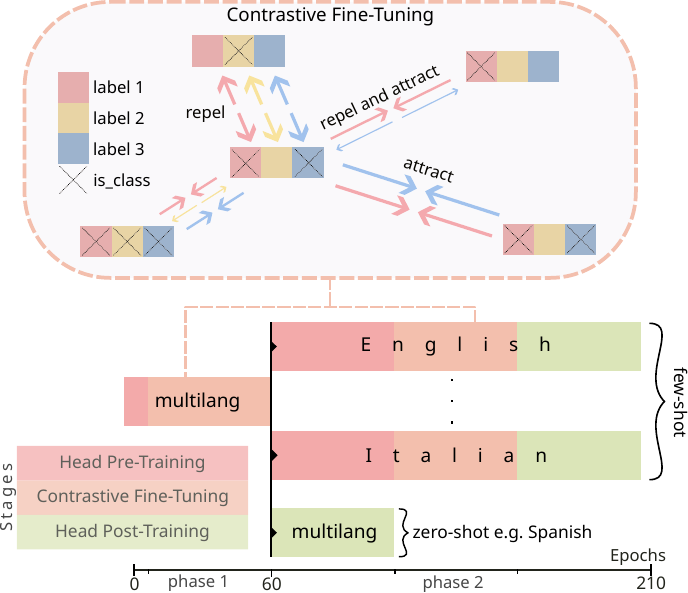}
    \caption{Our system performs label-aware contrastive fine-tuning (top). Embeddings of samples with similar labels are attracted, while they are repelled for dissimilar labels.
    The two-phase procedure (bottom) interleaves contrastive fine-tuning in both the multilingual and target language training.
    }
    \label{fig:training}
\end{figure}

We introduce \emph{mCPT}, the label-aware Contrastive Pre-training of Transformers based on a multilingual encoder model (original team name on the leaderboard\footnote{\url{https://propaganda.math.unipd.it/semeval2023task3/SemEval2023testleaderboard.html}}: \emph{PolarIce}). We exploit two features of the task: (i) multi-label information and (ii) multilingual data for pre-training. 

\begin{figure*}[t]
    \centering
    \begin{subfigure}[t]{0.27\textwidth}
        \centering
        \includegraphics[width=\textwidth]{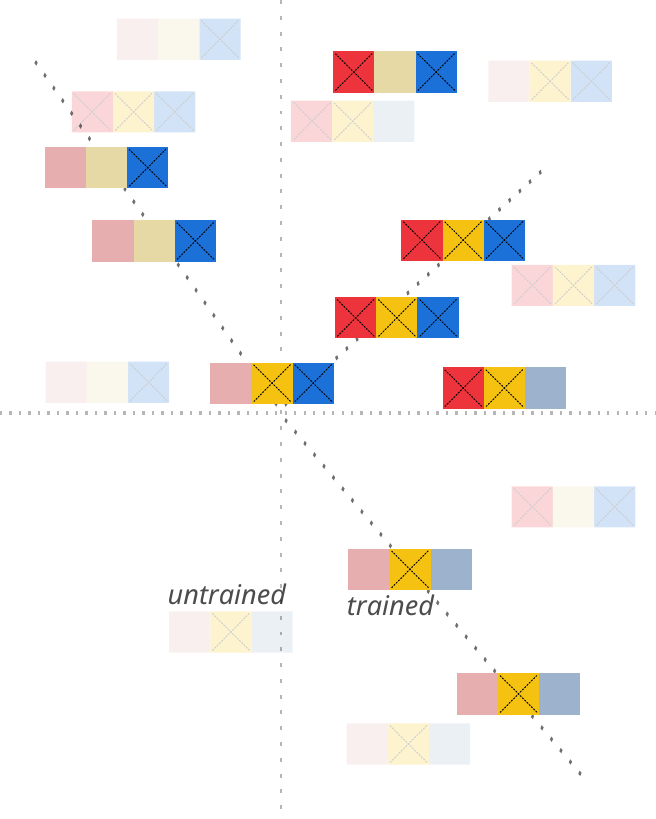}
        \caption{Repositioning due to loss}
        \label{fig:loss_fct}
    \end{subfigure}
    \hfill
    \begin{subfigure}[t]{0.37\textwidth}
         \centering
         \includegraphics[width=\textwidth,keepaspectratio]{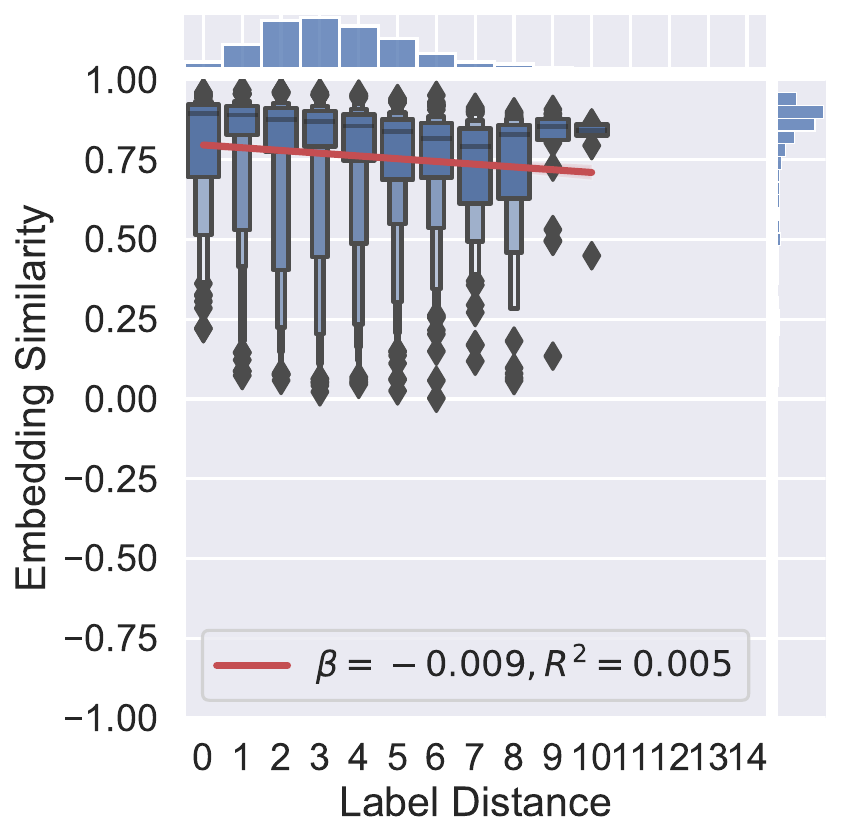}
         \caption{Before contrastive training %
         }
         \label{fig:similarity_before}
     \end{subfigure}
     \hfill
     \begin{subfigure}[t]{0.37\dimexpr\textwidth-1cm\relax}
         \centering
         \includegraphics[width=\textwidth,keepaspectratio,trim={1cm 0 0 0},clip]{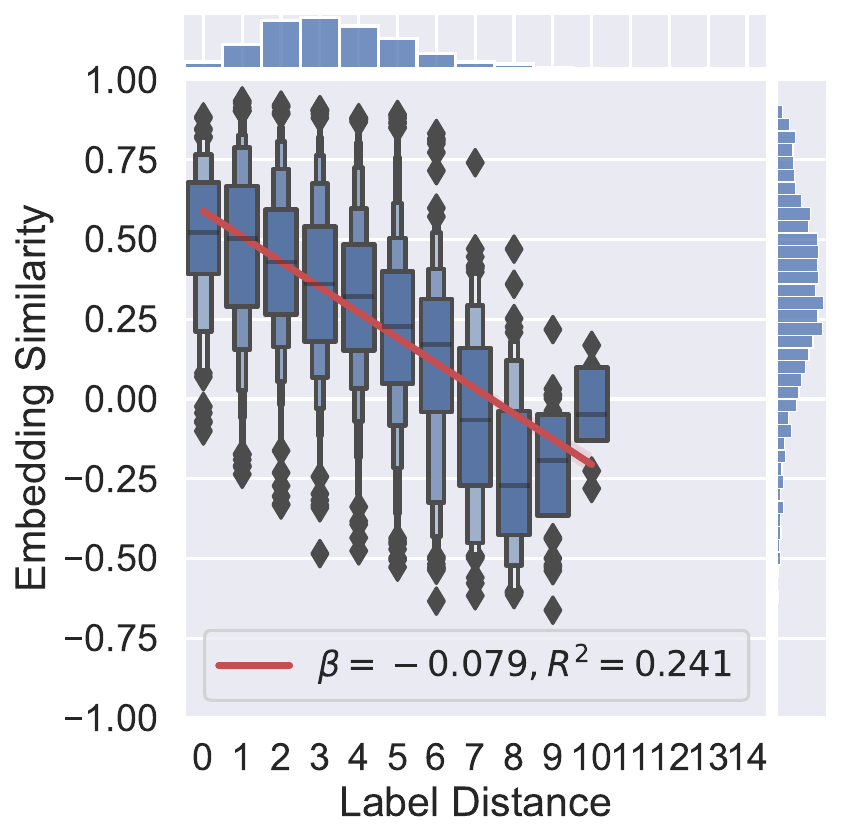}
         \caption{After contrastive training} %
         \label{fig:similarity_after}
     \end{subfigure}
    \caption{\emph{Effect of the loss function on the embedding space.} (a) Shows the repositioning of randomly generated samples (both embeddings and labels) in two-dimensional space. 
    The contrast loss function on its own increases the cosine similarity of latent representations with similar labels, while decreasing the similarity of representations with different labels. Note the positioning of trained embeddings with identical labels along lines drawn from the origin. (b) Without contrastive pre-training, the pairs of embeddings in the English dev set are similar regardless of their label distance. (c) After $50$ epochs, the embedding cosine similarity reflects the Hamming distance of the labels.
    }
    \label{fig:embedding_space}
\end{figure*}

First, we leverage the label information by adopting a contrastive loss function, i.e., HeroConLoss~\cite{zheng_heterogeneous_2022}, for natural language processing that optimizes the embedding space with respect to the similarities of the label space. Therefore, samples with more similar labels occupy similar regions in the embedding space, whereas mostly dissimilar samples regarding their shared labels are pushed apart (refer to Figures~\ref{fig:training} top and \ref{fig:loss_fct}). %

Second, we design a custom two-phase procedure with multiple stages for multilingual training to maximize the available data (see Figure~\ref{fig:training}). In phase one, we train on all languages, while in phase two, we further fine-tune the model on the target language if such data exist i.e.~few-shot setting, or continue training on all languages if not i.e.~zero-shot setting.

Our system performs competitively (top 10) on all six few-shot (i.e., English, German, French, Italian, Polish, and Russian) and three zero-shot (i.e., Spanish, Greek, and Georgian) settings, beating the baselines on all languages. On Spanish, which is the only zero-shot language with a common language family and alphabet as the training languages, our system is the winning contribution with a Micro-F1 of $0.571$ (compared to $0.120$ of the baseline). Therefore, we argue that our system generalizes well to unseen data, even when no training data is available in similar target languages.

In sum, our contribution is three-fold\footnote{Our code and model are publicly available at: \\\url{https://github.com/socialcomplab/semeval23-mcpt}}:

\begin{enumerate}[noitemsep,topsep=0pt]
    \item[C1] We adopt a multi-label contrastive loss function for natural language processing to optimize the embeddings of textual data.
    \item[C2] We describe a two-phase multi-stage training procedure for multilingual scenarios with limited data, i.e., few- and zero-shot predictions.
    \item[C3] We demonstrate the effectiveness of our winning system for framing detection supported by embedding and ablation studies.
\end{enumerate}

\section{Related Work}

\paragraph{Framing Detection.}
According to \citet{entman1993framing}, to frame is to select and emphasize some aspects of reality to encourage particular interpretations. That is, messages centered around a common topic may draw the receiver's attention to distinct features, thus suggesting different courses of action, causal interpretations, etc. As such, computational framing detection requires natural language processing (NLP) methods that capture nuances of \emph{how} content is presented rather than just \emph{what} topic is present. Therein, studies focus on detecting vastly different conceptualizations of framing, such as blame frames~\cite{shurafa2020political}, war frames~\cite{wicke2020framing}, moral frames~\cite{reiter2021studying}, or media frames~\cite{boydstun2014tracking,kwak2020systematic}. 

Regarding media frames, \citet{boydstun2013identifying} identified a set of relevant frames that formed the basis for the media frame corpus~\cite{card2015media}. Within this supervised frame detection scenario, \citet{liu2019detecting} indicate that Transformer-based approaches vastly outperform approaches using less powerful architectures such as LSTMs. 
As such, we also employ Transformer models with label-aware contrastive pre-training.

\paragraph{Supervised Contrastive Learning.}
Contrastive learning, originally mainly used in computer vision settings \cite[e.g.,][]{chopra2005learning}, has recently found increased attention in the NLP research community due to its efficacy on tasks with limited amounts of data and its applicability to Transformer embeddings~\cite[e.g.,][]{tunstall2022efficient}. The general concept of supervised contrastive learning~\cite{khosla2020supervised} is that latent representations (or embeddings in NLP) of samples with the same labels should be close in embedding space, while samples with different labels should be further apart.

\citet{su-etal-2022-contrastive} and \citet{zheng_heterogeneous_2022} have independently proposed contrastive learning methods for multi-label settings that weight similarities of samples by the similarity of their label vectors, i.e.~hidden representations of samples with similar label vectors should be more similar than hidden representations of samples with less similar label vectors. \citet{su-etal-2022-contrastive} weight a Euclidean distance-based measure of embeddings by a normalized dot product of the label vectors. HeroCon~\cite{zheng_heterogeneous_2022} generalizes supervised contrastive loss~\cite{khosla_supervised_2021} and beats previous state-of-the-art contrastive learning paradigms in multi-label settings on multiple image data sets. 
\citet{tunstall2022efficient} introduce SETFIT, an algorithm for the data-efficient fine-tuning of sentence embeddings, primarily on binary labels. SETFIT first fine-tunes sentence embeddings in a contrastive manner before training a classification head.

We combine the idea of the contrastive pre-training stage from
SETFIT and adopt HeroCon for NLP loss to improve performance on multi-label datasets.%

\begin{table*}[t]
    \centering
    \caption{\textbf{Test set results on the official leaderboard} on Subtask 2, first few-shot (top) then zero-shot (bottom). The results are sorted by Micro-F1 of \emph{mCPT}, i.e., our system performance on the target metric.
    Our system outperforms the \emph{Base} on all languages, both on Micro-F1 and Macro-F1, with the majority of improvements being very significant$^\dagger$. Similarly, \emph{mCPT} performs better than \emph{SETFIT} on all Latin-based languages. Our winning contribution to Spanish is also significantly better than SETFIT, as well as the averaged Micro and Macro-F1 scores.}
    \begin{threeparttable}
    \begin{tabular}{l | r | p{1cm} p{1cm} l | p{1cm} p{1cm} l | p{.1cm} c} %
        
         & \multicolumn{1}{c}{\# Samples} &  \multicolumn{3}{c}{\textbf{Micro-F1}} & \multicolumn{3}{c}{Macro-F1} & \multicolumn{2}{c}{Position}\\
        Language & Train/Dev/Test & \textbf{mCPT} & \textsc{SETFIT} & Base & \textbf{mCPT} & \textsc{SETFIT} & Base & \# & Teams \\
        \midrule

        German ($\mathcal{G}$, $L$)   & $132$ / \;$45$ / \;$50$ & \textbf{.622}$^{\ast}$ & $.549$  & $.487$ & \textbf{.564}$^{\ast}$ & $.492$   & $.418$ & 6 & /19 \\
        Polish ($\mathcal{S}$, $L$)   & $145$ / \;$49$ / \;$47$ & \textbf{.597} & $.584$   & $.594$ & \textbf{.555} & $.542$   & $.532$ & 9 & /19 \\
        Italian ($\mathcal{R}$, $L$)  & $227$ / \;$76$ / \;$61$ & \textbf{.584}$^{\ast}$  & $.502$  & $.486$ & \textbf{.469}$^{\ast\ast}$ & $.371$   & $.372$ & 5 & /19 \\
        English ($\mathcal{G}$, $L$)  & $433$ / \;$83$ / \;$54$ & \textbf{.535}$^{\ast}$  & $.469^{\ast}$   & $.350$ & \textbf{.482}$^{\ast}$ & $.409^{\ast}$   & $.274$ & 5 & /23 \\
        French ($\mathcal{R}$, $L$)   & 158 / \;53 / \;50 & \textbf{.469}$^{\ast}$ & $.463^{\ast}$   & $.329$ & \textbf{.429}$^{\ast}$ & $.419^{\ast}$   & $.276$ & 9 & /19 \\
        Russian ($\mathcal{S}$)  & $143$ / \;$48$ / \;$72$ & $.409^{\ast}$  & \textbf{.421}$^{\ast}$  & $.230$ & \textbf{.367}$^{\ast\ast}$ & $.258$   & $.218$ & 5 & /18 \\
        \midrule
        \rowcolor[gray]{.9} \emph{Spanish} ($\mathcal{R}$, $L$)  &   $-$ / \;\;$-$ / \;$30$ & \textbf{.571}$^{\ast\ast}$  & $.418^{\ast}$  & $.120$ & \textbf{.455}$^{\ast\ast}$ & $.305^{\ast}$   & $.095$ & \underline{\textbf{1}} & /17 \\
        \emph{Greek}    &   $-$ /  \;\;$-$ / \;$64$ & \textbf{.516}$^{\ast}$ & $.427$   & $.345$ & \textbf{.410}$^{\ast}$ & $.338^{\ast}$   & $.057$ & 7 & /16 \\
        \emph{Georgian} &  $-$ / \;\;$-$ / \;$29$ & $.400^{\ast}$ & \textbf{.404}$^{\ast}$    & $.260$ & $.291$ & \textbf{.384}$^{\ast}$   & $.251$ & 9 & /16 \\ \midrule \midrule
        Summary & $1238$ /$354$ /$457$ & \textbf{.523}$^{\ast\ast}$ & $.471^{\ast}$ & $.356$ & \textbf{.447}$^{\ast\ast}$ & $.391^{\ast}$ & $.277$ & $6.\overline{2}$ & /$18.\overline{4}$\\
        \bottomrule
    \end{tabular}
    \begin{tablenotes}
    \footnotesize
    \item[mCPT] Our system; \item[SF] \textsc{SETFIT} Transformer model; \item[Base] Challenge Baseline (n-grams count + SVC); \\
    \item[$\dagger$] We assume a normal approximation interval on a binomial distribution 99.5\% confidence level ($z=2.81$)
    concerning the number of labels as proxy. We will update the table with a statistical test on the samples once the test labels are released. \\
    \item[$\ast$] Significant improvement outside the confidence interval compared to Base; \item[$\ast\ast$] also over \textsc{SETFIT}; \item[\underline{\textbf{1}}] Winner; \\
    \item[\textbf{{Bold}}] Best performance; \item[\emph{Italic}] Zero-Shot Language; 
    \item[$\mathcal{G}$] Germanic; \item[$\mathcal{S}$] Slavic; \item[$\mathcal{R}$] Romance; \item[$L$] Latin alphabet; 
    \end{tablenotes}
    \end{threeparttable}
    \label{tab:test_results}
\end{table*}

\section{Methods}
At the core of our system lies a multilingual Transformer model with dense neural layers comprising the head. Contrastive fine-tuning is performed as part of a multi-stage training procedure. 

\subsection{Contrastive Fine-Tuning} \label{sec:contrastive_fine_tuning}
Our contrastive fine-tuning objective (\emph{C1}, Figure~\ref{fig:training} top) is centered around the idea that embeddings of samples with similar labels should be close while embeddings of samples with very distinct label vectors should be distant. Following \citet{zheng_heterogeneous_2022} for every batch and every class, we compute the similarity between positive samples, i.e., samples that are of that class, and all others. As such, samples may both repel and attract each other within different classes yet do neither if they are both negative.

Our loss function is a linear combination of two terms: 
A binary cross entropy term $\mathcal{L}_{BCE}$ that jointly optimizes the head and body
in the contrastive fine-tuning stage and a contrastive term $\mathcal{L}_{CON}$: 
\begin{equation} \label{eq:total_loss}
    \mathcal{L} = \mathcal{L}_{BCE} + \alpha \mathcal{L}_{CON}
\end{equation}
where $\alpha$ is a weighting hyperparameter.
The contrastive loss is given by:
\begin{equation} \label{eq:contrast_loss}
\resizebox{.84\hsize}{!}{$
    \mathcal{L}_{CON} = \frac{1}{|C|} \underset{c\in C}{\sum} -\mathbf{E}_{X_i, X_j \in \mathcal{P}(c)} \left[ \log \frac{\sigma_{ij} f(X_i, X_j)}{\delta_{ij}} \right]
$}
\end{equation}
 where $C$ is the set of all classes (e.g.~\emph{Economic}), $\mathcal{P}(c)$ is the set of all positive samples i.e., all embeddings $X_i$ that are of class $c$, and $f(\cdot, \cdot)$ is the cosine similarity measure between embeddings. The loss is normalized by:
\begin{equation}\label{eq:contrast_normalize}
\resizebox{.84\hsize}{!}{$
    \delta_{ij} = \frac{\sigma_{ij} f(X_i, X_j) + \sum_{X_k \in \mathcal{N}(c)} \gamma_{ik} f(X_i, X_k)}{|\mathcal{N}(c)| + 1}
$}
\end{equation}
where $\mathcal{N}(c)$ is the set of all negative samples and $\sigma_{ij}$ and $\gamma_{ik}$ are given by:
\[\sigma_{ij} = 1 - d(Y_i, Y_j) / |C|,  \quad \gamma_{ik} = d(Y_i, Y_k)\]
respectively where $d$ describes the Hamming distance between label vectors $Y_i$.

\subsection{Training Procedure}

We follow a two-phase training procedure illustrated in Figure~\ref{fig:training} (bottom), focusing on optimally utilizing the available data, which in our case are the six languages of subtask 2 (\emph{C2}). Inspired by \citet{tunstall2022efficient}, we first optimize the embedding space of a (multilingual) Transformer model.
Herein, our approach makes the assumption that the embeddings, regardless of language, possess mutual information given similar labels~\cite{zheng_heterogeneous_2022}. 
While the embedding space may be improved in this manner, we fine-tune on the target language to further improve the performance.

Phase one consists of two separate stages: \emph{head pre-training} and \emph{contrastive fine-tuning}. 
For phase two, the head is re-used for zero-shot settings, while discarded and randomly re-initialized for few-shot settings. In the former case, we conduct the \textit{post-training} stage on all languages, while in the latter it is essential to \emph{pre-train} the head on the target language before proceeding to the \textit{contrastive fine-tuning} and \textit{head post-training} stages. 
Both the \emph{head pre-training} and \emph{head post-training} stages only compute the binary cross entropy term $\mathcal{L}_{BCE}$, while simultaneously leaving the body unchanged, i.e., frozen. 
While the stages are identical, the rationale for each of them is very distinct:
As the head is randomly initialized, we first \emph{pre-train} it to avoid high gradients in the subsequent \emph{contrastive fine-tuning} stage. In contrast, \emph{post-training} allows the head to better fit the fine-tuned embeddings.

\section{Experiments}
We now present the results of our \emph{mCPT} system, supported by embedding space and ablation studies~(\emph{C3}).
We base \emph{mCPT} on the
multilingual\footnote{The base model was trained on 50+ languages including all nine of the shared task, thus being suitable for the problem.} sentence Transformer model \textit{paraphrase-multilingual-MiniLM-L12-v2}~\cite{reimers-2019-sentence-bert} and demonstrate that competitive results can be achieved with a relatively small amount of parameters, i.e., $117M$ parameters~\cite{wang2020minilm},  given a training method tailored to the task. The model was chosen for its sentence embedding performance on multiple languages and its small size relative to similar state-of-the-art multilingual Transformer models.

Our model architecture comprises mean-pooling, no normalization of embeddings, a dense head with one hidden layer of size $256$, and a dropout of $0.5$. 
We train the model with separate 
learning rates for the classification head ($1e-3$) and the body ($2e-5$), a 
weighting parameter $\alpha$ of $0.01$, a batch size of $26$ for $10$ and $50$ epochs for \emph{head pre-training} and \emph{contrastive fine-tuning} respectively in phase one (more details in Appendix~\ref{app:training_env}). 

\paragraph{Baseline Models.}
We compare the performance of our system against two baselines. First, we consider the results of the official baseline \emph{Base}~\cite[i.e., n-grams and support vector classification; ][]{semeval2023task3}. 
Second, we compare against SETFIT~\cite{tunstall2022efficient} with the same base encoder as ours, on the post-challenge test set (details in Appendix~\ref{app:setfit_params}).

\subsection{Main Results}

\emph{mCPT} performs better on Latin alphabets (marked by $L$) in both few- and zero-shot settings, and improves upon the two baselines (as presented in Table~\ref{tab:test_results}). 
In Slavic languages ($\mathcal{S}$), Polish is second-best in terms of Micro-F1 ($0.597$), but only slightly outperforms the baseline, whereas the improvement on Russian is very significant but only achieves the second-lowest score of $0.409$. In comparison, both Germanic languages ($\mathcal{G}$), German and English perform well, where we also have our highest overall Micro-F1 of $0.622$ for German, but also a high baseline of $0.487$. 
Although we find significant improvement on Greek and Georgian (which are zero-shot languages that do not share a major branch with any other language) over the baseline, both perform poorly in terms of Micro-F1 (i.e., Georgian having the lowest Micro-F1 of $0.400$). Hence, we suspect that not enough information from the other languages could be transferred. Finally, in the Romance languages ($\mathcal{R}$), Italian and Spanish perform well, while French with a Micro-F1 of $0.516$ performs lower than Greek.

The performance of Spanish is especially noteworthy, as it is the only zero-shot language that shares an alphabet as well as a language family with the training data languages. Therefore, we argue that our winning performance on Spanish (Micro-F1 of $0.571$ compared to $0.12$ of the baseline) stems from the fact that the knowledge was successfully transferred from the other languages to the zero-shot setting. This is further supported by \emph{SETFIT}, which improves upon the baseline but shows lower performance on Latin-based languages.%

\subsection{Embedding Space Analysis}
\label{sec:emb_analysis}

Figure~\ref{fig:embedding_space} demonstrates how our contrastive training procedure optimizes the embeddings of the Transformer body. Figure~\ref{fig:loss_fct} exemplarily shows the repositioning of samples due to the loss function in two-dimensional space. Observe how the trained labels (opaque) align, with the blue-yellow label between the two blue-only labels and two yellow-only labels, while simultaneously pushing the red labels to the side (more detailed analysis in Appendix~\ref{app:toy_example}).
For the analysis of the high-dimensional embeddings ($384$) and label spaces ($14$) on the dev set, we use boxen plots concerning embedding cosine similarity for all pairwise samples within a given Hamming distance.

Regarding the pre-trained base model on English (Figure~\ref{fig:similarity_before}),
we find a suboptimal correlation with $R^2=0.005$ and $\beta=-0.009$. 
In comparison, after contrastively training the model, the correlation becomes much more pronounced, i.e., $R^2=0.241$ and $\beta=-0.079$ for English as shown in Figure~\ref{fig:similarity_after}. Furthermore, the spread of pairwise embedding similarity distribution increases, as a greater amount of samples become dissimilar to each other, especially for higher label distance. Thus, we conclude that our system leads to higher utilization of the available embedding space, which in turn boosts performance. Appendix~\ref{app:label_correlations} contains the remaining languages. %

Finally, we want to emphasize that our data set has no perfectly dissimilar label vector pairs, which would make hard negative mining approaches~\cite{gao2021simcse} infeasible, e.g., for using plain contrastive loss~\cite{chopra2005learning}.

\begin{table}[t]
    \centering
    \caption{Ablation study (top) and the proposed contrastive sampling extension (bottom) on the dev set. In general, we observe that \emph{mCPT} performs best with all components, i.e., pre-training (PT), contrastive loss ($\mathcal{L}_{CON}$), and end-to-end training (E2E), enabled. Contrast sampling (CS) suggest further improvements.
    }
    
    \setlength\tabcolsep{4pt} 
    \begin{tabular}{l c c c c c c } \toprule
        Model & en & it & ru & fr & ge & po  \\ \midrule
        mCPT & \textbf{.682} &  \textbf{.585} & \textbf{.520} & \textbf{.570} & .561 & .636 \\
        - PT & .681 & .545 & .475 & .563 & .583 & .616 \\
        - $\mathcal{L}_{CON}$ & .657 & .521 & .436 & .524 & .570 & \textbf{.645} \\
        - E2E & .629 & .519 & .500 & .535 & \textbf{.586} & .633 \\ \midrule
        mCPT+CS & \textbf{.688} & \textbf{.590} & .519 & \textbf{.575} & \textbf{.591} & .638 \\ 
        \bottomrule
    \end{tabular}
    \label{tab:ablation}
\end{table}

\subsection{Ablation Study and Extension}
Table~\ref{tab:ablation} indicates the effectiveness of our combined training approach \emph{mCPT}. From \emph{mCPT}, we remove components iteratively, first removing the multilingual pre-training phase, then the contrastive term (see Equation~\ref{eq:contrast_loss}), and finally, end-to-end training leaving only a trained classification head with no embedding fine-tuning. Comparing the results of the ablation study with those of Table~\ref{tab:test_results} it is interesting to note that our approach works best on languages with lower scores. A hypothesis is that the out-of-the-box Transformer embeddings already fit the data well and that \emph{mCPT} is not able to improve upon the already strong baseline.
Finally, we find that adding a contrast sampling extension could further improve the results (see Appendix~\ref{app:contrast_sampler}).

\section{Conclusion}

In this paper, we describe our system (\emph{mCPT}) for the framing detection shared task~\cite{semeval2023task3}. 
We introduce an approach based on a label-aware contrastive loss and training procedure for Transformers to deal with the challenges of multilingual multi-label prediction with few or even zero samples. The generalization ability of our system is demonstrated by providing the winning contribution for the Spanish framing detection subtask where no training samples were available\footnote{We refer to Appendices~\ref{sec:limitations} and \ref{sec:ethics} for discussions on limitations and ethical considerations, respectively.}. Hence, we believe that our system is a notable advancement for computational framing research.

\section*{Acknowledgements}

We would like to thank the anonymous reviewers for taking the necessary time and effort to review the manuscript and appreciate their universal positive feedback.

\bibliography{anthology,custom}
\bibliographystyle{acl_natbib}

\appendix

\section{Training Environment}
\label{app:training_env}

We performed the main experiments on the Kaggle platform (\url{www.kaggle.com}) with the P100 graphics card. We chose a free platform for the computation to demonstrate that our system is tailored towards the task at hand and is accessible for everybody, rather than relying on large amounts of computational resources. We empirically selected the hyperparameters to fit the platform. Herein, we chose a batch size of $26$ which optimally utilizes the available GPU memory. The multilingual pre-training takes approximately 1.5 hours, while the language-specific fine-tuning takes 1 hour each.

\section{SETFIT Parameters}
\label{app:setfit_params}

We choose \emph{SETFIT} as it is similar in concept to our system, i.e., contrastive learning for Transformers, but not aligned with the shared task, i.e., does not explicitly consider multi-label problems. 
Hence, the comparison demonstrates how our system is an improvement over established approaches in this setting and emphasizes that the adaptions of the contrastive loss and training procedure are indeed beneficial. We report the results of \emph{SETFIT} on the post-challenge leaderboard without further adaption after the initial submissions for fair comparisons on the test set.

We mimic the parameters setting where applicable while preserving the standard training procedure to maximize comparability. 
We first contrastively train the body for $10$ epochs before training the full model end-to-end for $50$ epochs with a batch size of $26$ with learning rates of $1e-3$ and $2e-5$ for the head and the body respectively. The body-then-end-to-end procedure was suggested by the usage guide. The training runtime is approximately $10$ hours on the Kaggle platform (which again was chosen for a fair comparison). Initially, we experimented with SETFIT in the challenge period but decided to submit our presented system instead.

\section{Repositioning of Samples}
\label{app:toy_example}

In Figure~\ref{fig:loss_fct}, we show the effect of the loss function and how the label and embedding space are intertwined. Specifically, we demonstrate how randomly generated embeddings in two-dimensional space with three-dimensional label vectors shift towards more optimal positions after applying the contrastive loss function. Accordingly, the initially random positions of embeddings with equivalent labels end up on straight lines drawn from the origin. The observed effect is a direct consequence of similar label vectors attracting and opposite labels repelling each other. Moreover, a partial label similarity with two distinct groups ends between those groups, as a result of both forces being active. For instance, consider the line from top left to bottom right: blue-only labels become attracted, and repel yellow-only labels, while the sample with blue and yellow lies in between. Due to the resulting positioning, the embedding space becomes disentangled leading to an increase in linear separability, which benefits classifiers such as our differentiable head.

\begin{figure}[!pht]
    \centering
    \begin{subfigure}[t]{0.245\textwidth}
         \centering
         \includegraphics[width=\textwidth,keepaspectratio]{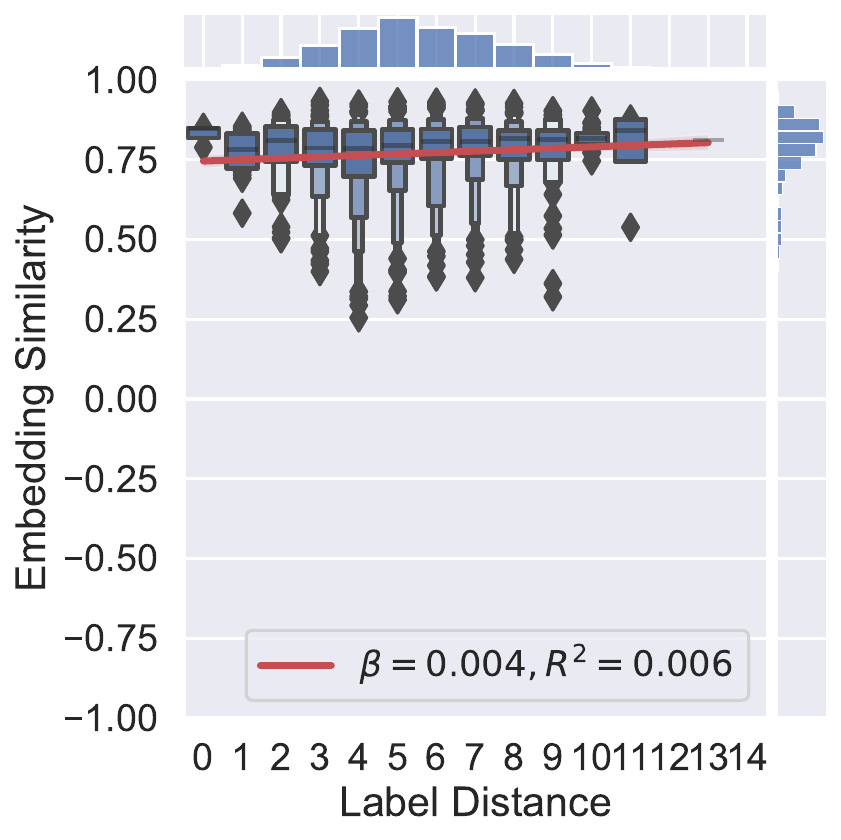}
         \caption{Before on German.}
         \label{fig:similarity_before_ge}
    \end{subfigure}
    \hfill
    \begin{subfigure}[t]{0.245\dimexpr\textwidth-1cm\relax}
        \centering
        \includegraphics[width=\textwidth,keepaspectratio,trim={1cm 0 0 0},clip]{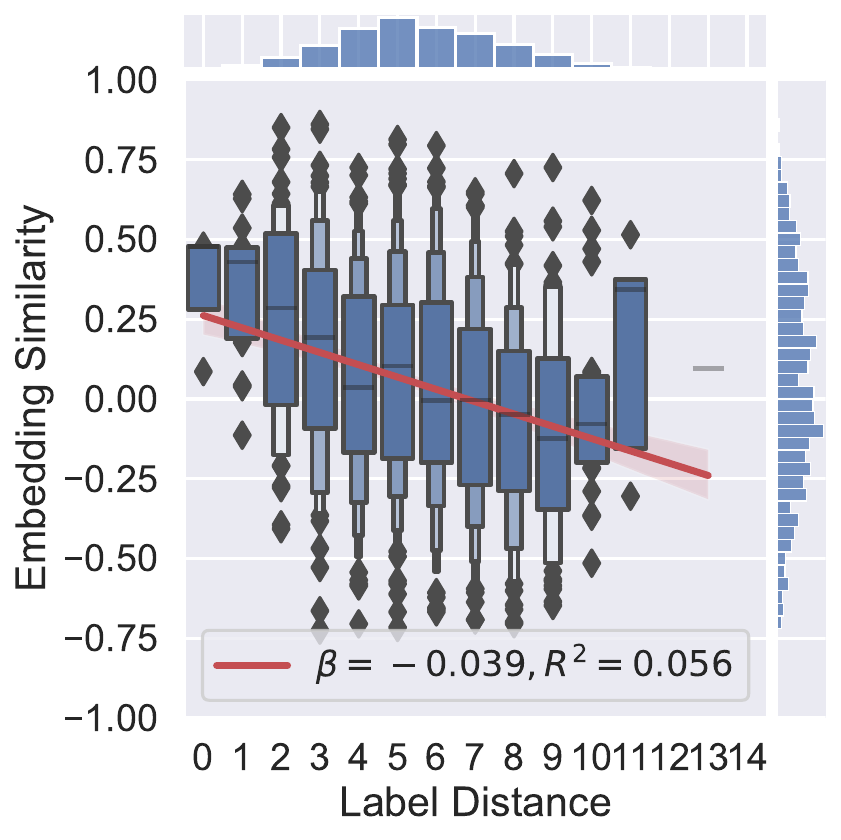}
        \caption{After on German.}
        \label{fig:similarity_after_ge}
    \end{subfigure}
    \hfill
    \begin{subfigure}[t]{0.245\textwidth}
         \centering
         \includegraphics[width=\textwidth,keepaspectratio]{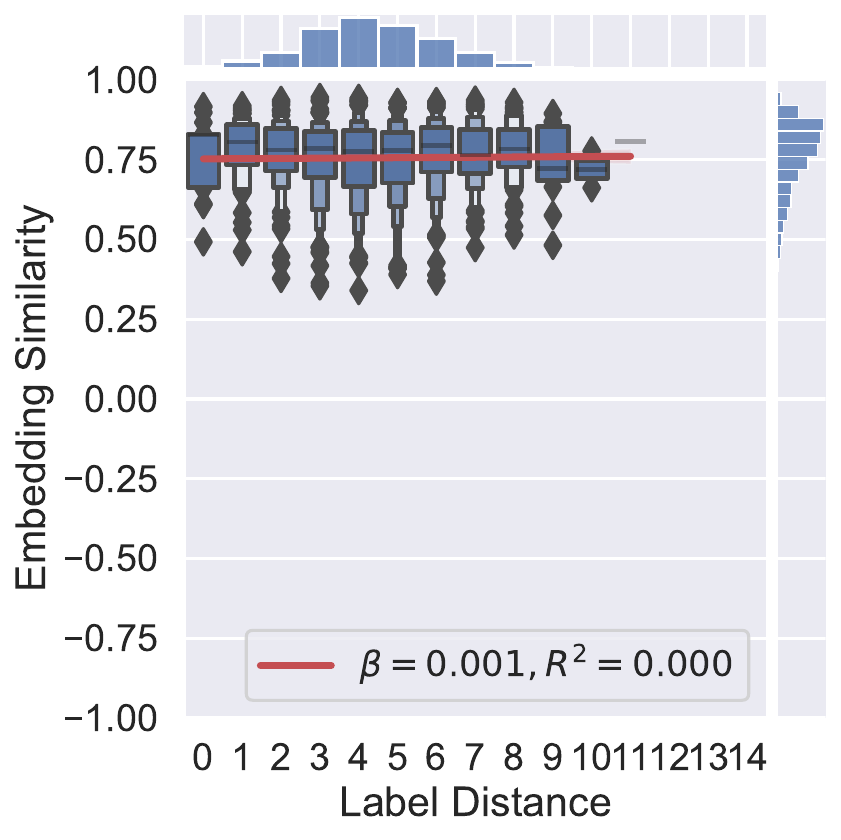}
         \caption{Before on French.}
         \label{fig:similarity_before_fr}
    \end{subfigure}
    \hfill
    \begin{subfigure}[t]{0.245\dimexpr\textwidth-1cm\relax}
        \centering
        \includegraphics[width=\textwidth,keepaspectratio,trim={1cm 0 0 0},clip]{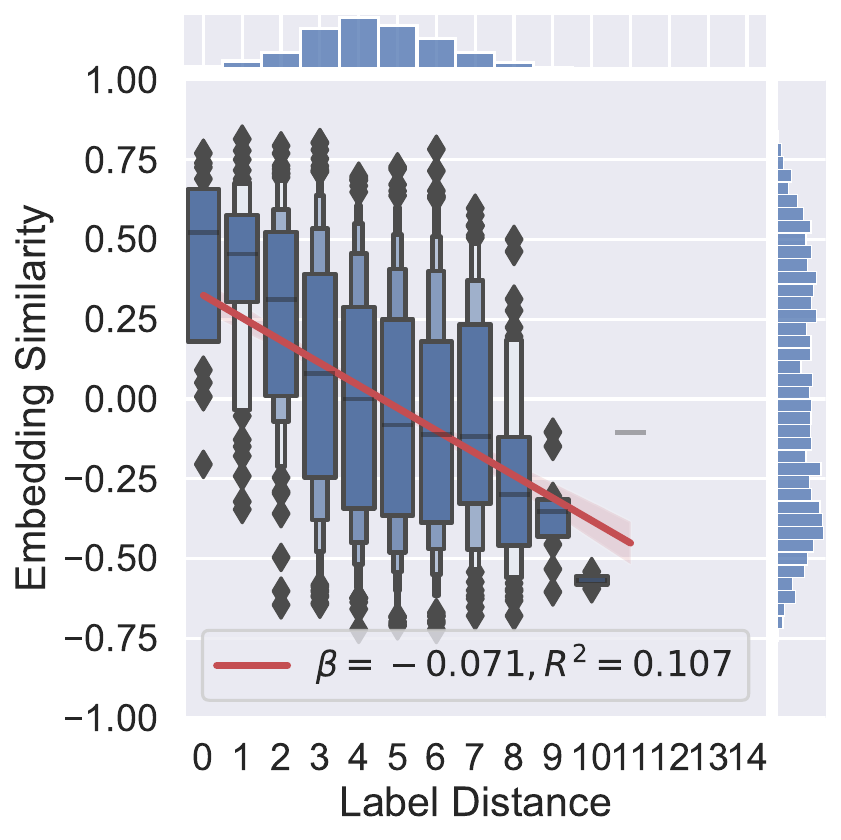}
        \caption{After on French.}
        \label{fig:similarity_after_fr}
    \end{subfigure}
    \hfill
    \begin{subfigure}[t]{0.245\textwidth}
         \centering
         \includegraphics[width=\textwidth,keepaspectratio]{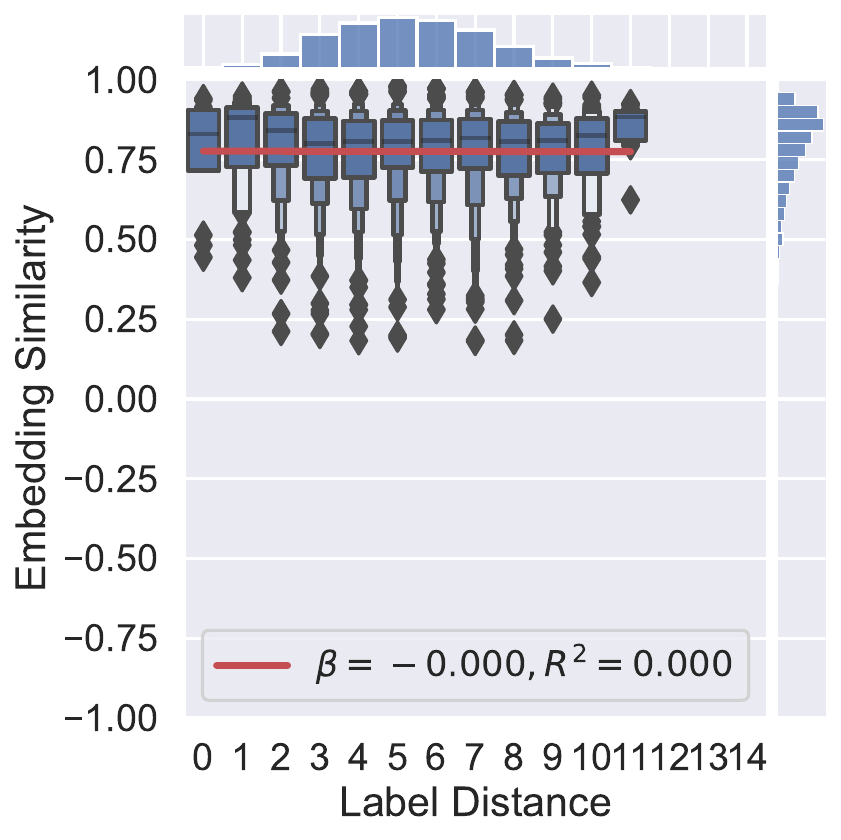}
         \caption{Before on Italian.}
         \label{fig:similarity_before_it}
    \end{subfigure}
    \hfill
    \begin{subfigure}[t]{0.245\dimexpr\textwidth-1cm\relax}
        \centering
        \includegraphics[width=\textwidth,keepaspectratio,trim={1cm 0 0 0},clip]{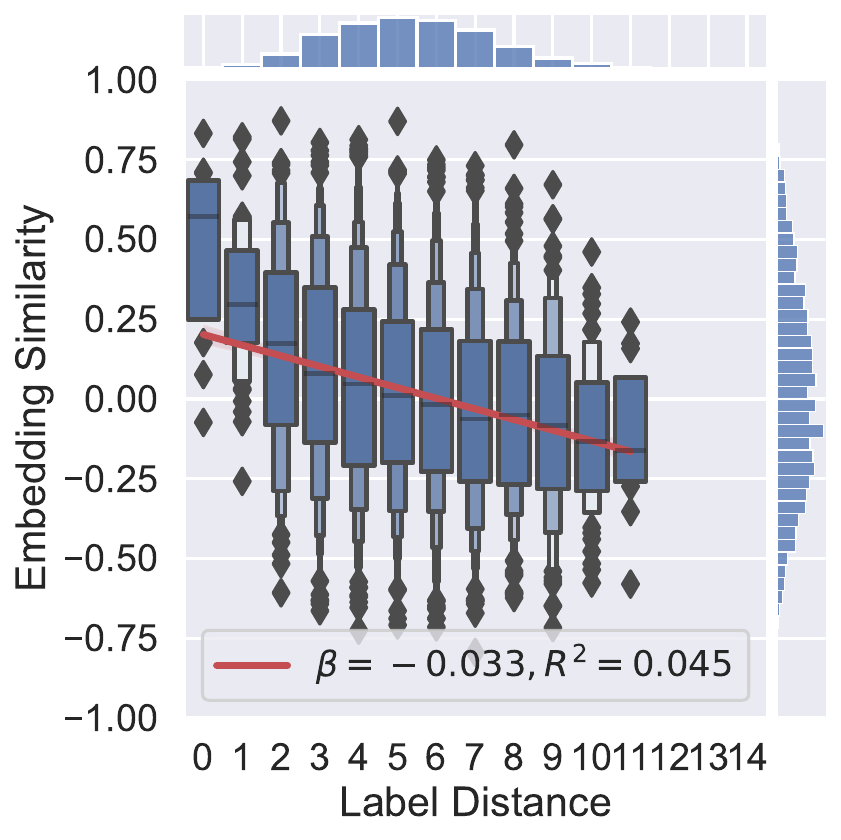}
        \caption{After on Italian.}
        \label{fig:similarity_after_it}
    \end{subfigure}
    \hfill
    \begin{subfigure}[t]{0.245\textwidth}
         \centering
         \includegraphics[width=\textwidth,keepaspectratio]{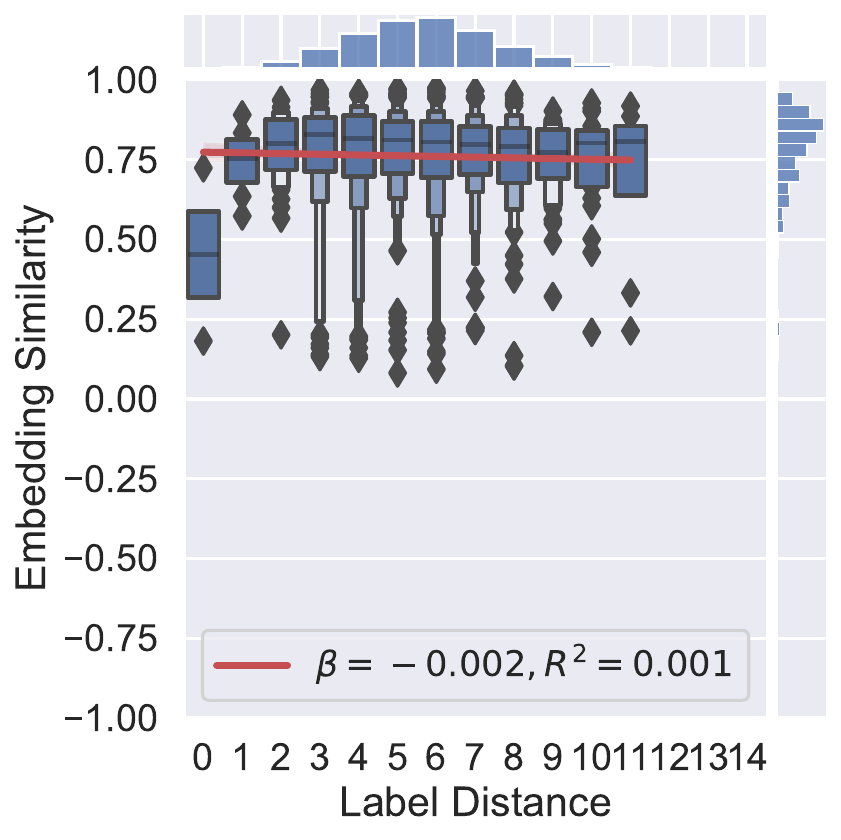}
         \caption{Before on Polish.}
         \label{fig:similarity_before_po}
    \end{subfigure}
    \hfill
    \begin{subfigure}[t]{0.245\dimexpr\textwidth-1cm\relax}
        \centering
        \includegraphics[width=\textwidth,keepaspectratio,trim={1cm 0 0 0},clip]{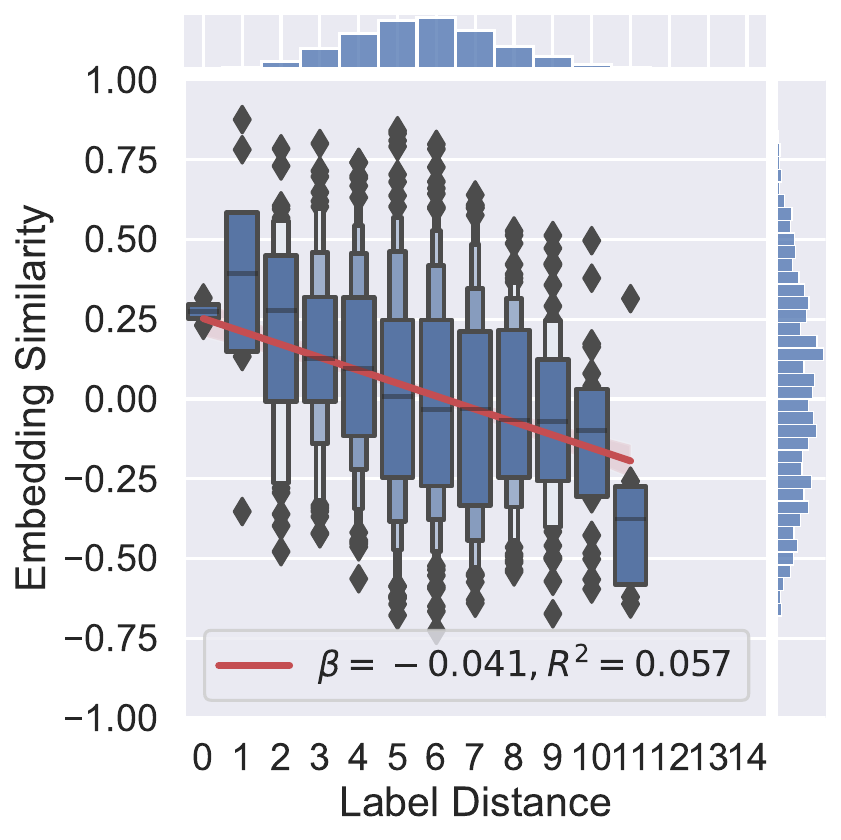}
        \caption{After on Polish.}
        \label{fig:similarity_after_po}
    \end{subfigure}
    \hfill
    \begin{subfigure}[t]{0.24\textwidth}
         \centering
         \includegraphics[width=\textwidth,keepaspectratio]{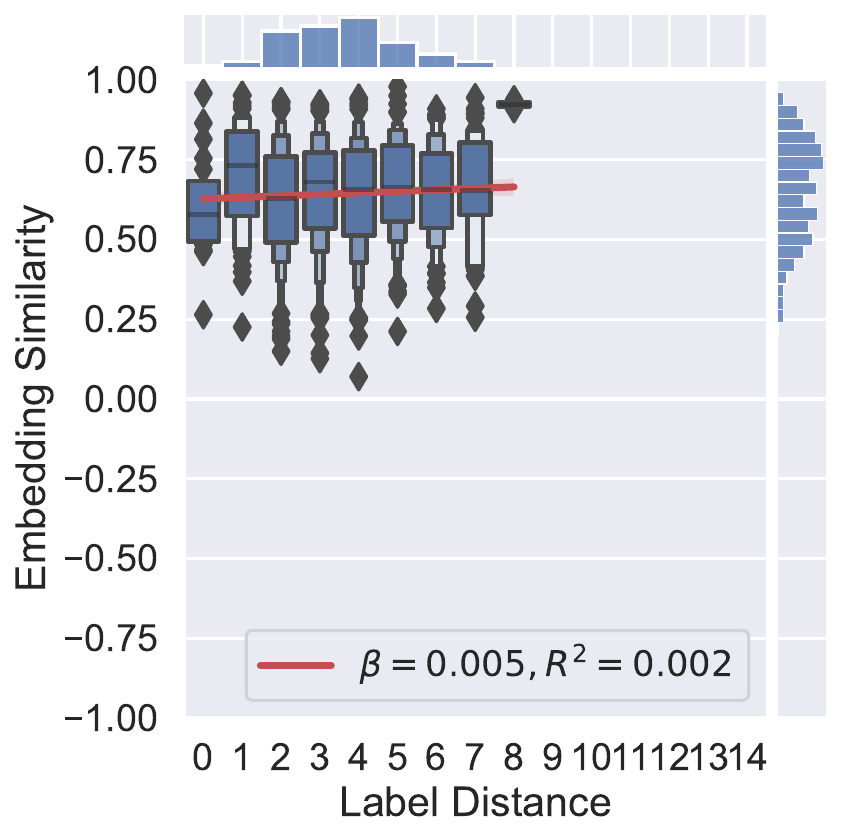}
         \caption{Before on Russian.}
         \label{fig:similarity_before_ru}
    \end{subfigure}
    \hfill
    \begin{subfigure}[t]{0.245\dimexpr\textwidth-1cm\relax}
        \centering
        \includegraphics[width=\textwidth,keepaspectratio,trim={1cm 0 0 0},clip]{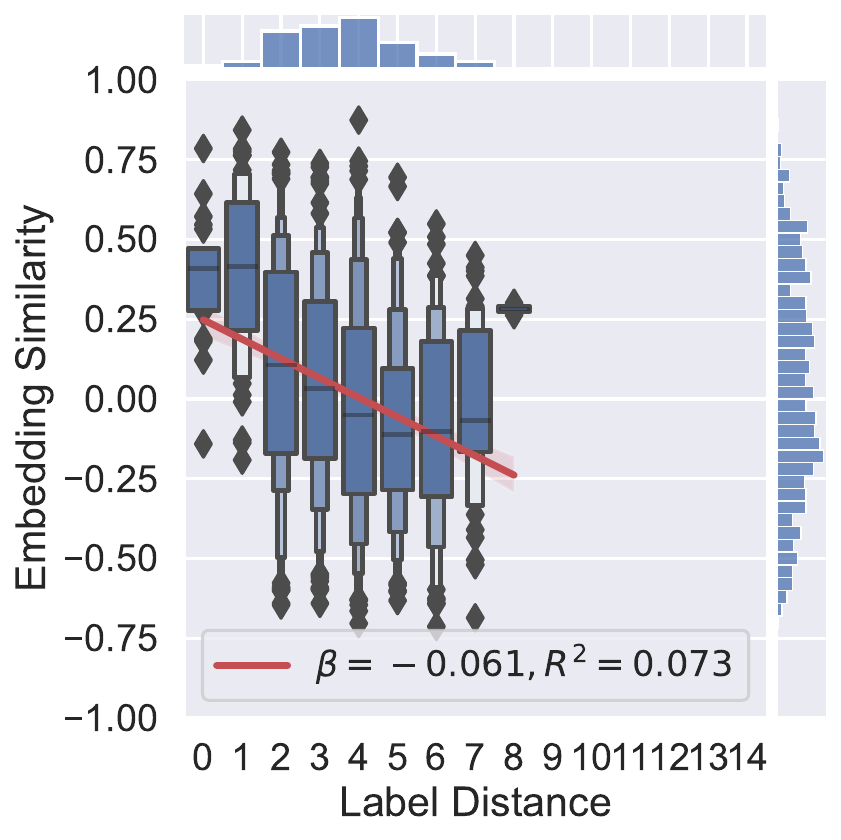}
        \caption{After on Russian.}
        \label{fig:similarity_after_ru}
    \end{subfigure}
    \caption{\emph{Effect of the loss function on the embedding space.} Evaluated on various language dev sets.
    }
    \label{fig:embedding_space2}
\end{figure}

\section{Embeddings and Labels Correlation}
\label{app:label_correlations}

Here, we present the remaining languages for the embedding space analysis in Figure~\ref{fig:embedding_space2}. 
When considering the plots before contrastive training is performed, the correlations between embedding spaces and label spaces are very weak, as well as most embeddings being very similar regardless of their label distance. Hence, virtually no pair of samples are opposite to each other, i.e., has a noteworthy negative cosine similarity. Conversely, all pairs are similar to a certain extent.
We argue that in these given embedding spaces, it is challenging for a classifier to learn a discriminative function. Moreover, due to the suboptimal positioning of embeddings, a substantial amount of the embedding space does not contribute to the prediction performance, thus wasting the model's potential expressiveness.
Hence, the interpretation from Section~\ref{sec:emb_analysis} can be directly applied to the five other languages, as the effect is the same (although differently pronounced). 
For instance, models of German and Russian even have a slight upward slope when applied without contrastive training. Hence, the negative slope and regression fit increases with contrastive training, as intended and expected.

\section{Contrast Sampling Extension.}
\label{app:contrast_sampler}
Adding a custom \emph{contrast sampler} which ensures that at least one sample from every class is present per batch further improves consistency as well as performance. Due to the nature of the contrastive objective, it is imperative that every batch contain negative as well as positive pairs of samples for every class. This is not guaranteed by sampling randomly, especially if the label distribution is imbalanced. As illustrated by Table~\ref{tab:ablation} it outperforms pure \emph{mCPT} in five out of six languages while coming in second by a small margin in Russian. We attribute this largely to the variance introduced by using a relatively small batch size of 26 compared to the number of labels (14). This  variance may lead to undesirable gradient updates in some iterations when batches contain label distributions that are not representative. %

\section{Limitations}
\label{sec:limitations}

We recognize three main limitations of our work, which are distinct in their aspect. 

First, the \emph{performance limitation}; while our system has competitive results across the board, it only performs best in one of the nine languages on the leaderboard. In comparison, team \emph{MarsEclipse}~\cite{MarsEclipseSemeval2023task3}, which also focused on the framing detection subtask, wins all six few-shot languages and performs second on two of three (i.e., Greek and Georgian) zero-shot languages. They only perform worse at Spanish (6\textsuperscript{th}), which is opposite to our placement. Team \emph{SheffieldVeraAI}~\cite{SheffieldVeraAISemeval2023task3}, who also participated in the other two subtasks regarding news genre and persuasion technique detection, perform well across the board and wins the Greek and Georgian framing detection tasks. Hence, our system occupies the niche of zero-shot prediction when trained with similar languages (i.e, in our case Latin-based). 

Second the \emph{technical limitation}, our system was trained using a small multilingual model as we aimed towards adapting Transformer pre-training for the multi-label challenge in particular rather than achieving the highest performance with computationally expensive training. However, as a consequence, we do not know how well our system scales to bigger models, such as MPNet~\cite{song2020mpnet}, and plan to address this limitation in future work.

Third the \emph{task setting limitation}, we want to emphasize a potential limitation resulting from the shared task setting. \citet{ali-hassan-2022-survey} argue that the specified labels in the media frame corpus~\cite{card2015media} revolve around topics (i.e., the \emph{what}) rather frames (i.e., the \emph{how}). As the same labels were adopted for the shared task, the conceptualizations of frames are expected to be similar to a certain extent. They thus would also affect the resulting models and predictions.

\section{Ethics Statement}
\label{sec:ethics}

We want to discuss three ethical considerations of our system. First, our system is based on pre-trained Transformers, which inherit \emph{biases} from their training data. For the shared task, these biases are negligible, but are a concern for real-world applications. The second consideration relates to \emph{fairness} concerns. The performance varies strongly between languages, with more researched languages typically resulting in better performance. We, thus, embrace the multilingual setting of the shared task with one-third zero-shot languages, but similarly achieved better performance in Latin-based languages. Third, our system leads to better detection of media frames, which is an important research direction. However, the system could in theory also be used in a disputed or even malicious manner, e.g., for \emph{reframing} political statements. Hence, we do not advise specific applications of our system besides better framing detection.

\end{document}